\documentclass[journal]{IEEEtran}

\ifCLASSINFOpdf
\else
   \usepackage[dvips]{graphicx}
\fi
\usepackage{url}

\usepackage{amsmath,graphicx,setspace}
\usepackage{amssymb}
\usepackage{algorithm}
\usepackage{algorithmic}
\usepackage{textcomp}
\usepackage{subfigure}
\usepackage{tipa}
\usepackage{cite} 
\usepackage{url}
\usepackage{float}
\usepackage{multirow}

\makeatletter 
\renewcommand{\@thesubfigure}{\hskip\subfiglabelskip}
\makeatother

\begin{document}

\title{Dataset Distillation Using Parameter Pruning}

\author{Guang Li, \IEEEmembership{Graduate Student Member, IEEE}, Ren Togo, \IEEEmembership{Member, IEEE}, Takahiro Ogawa, and\\ Miki Haseyama, \IEEEmembership{Senior Member, IEEE}
\thanks{
This research was supported in part by the Hokkaido University-Hitachi Collaborative Education and Research Support Program and AMED Grant Number JP22zf0127004h0002.
All experiments were conducted on the Data Science Computing System of Education and Research Center for Mathematical and Data Science, Hokkaido University.
}
}

\markboth{Journal of \LaTeX\ Class Files, Vol. 14, No. 8, August 2015}
{Shell \MakeLowercase{\textit{et al.}}: Bare Demo of IEEEtran.cls for IEEE Journals}
\maketitle

\begin{abstract}
In this study, we propose a novel dataset distillation method based on parameter pruning.
The proposed method can synthesize more robust distilled datasets and improve distillation performance by pruning difficult-to-match parameters during the distillation process.
Experimental results on two benchmark datasets show the superiority of the proposed method.
\end{abstract}


\IEEEpeerreviewmaketitle

\section{Introduction}
\IEEEPARstart{L}{arge} datasets containing millions of samples have become the standard for obtaining advanced models in many artificial intelligence areas, including natural language processing, speech recognition, and computer vision~\cite{liu2017survey}.
Meanwhile, large datasets also raise some issues.
For example, data storage and preprocessing are becoming increasingly difficult.
Furthermore, expensive servers are required to train models on these datasets, which is not friendly for low-resource environments~\cite{xiong2022ld}.
An effective way to solve these problems is data selection, which identifies representative training samples of large datasets~\cite{bachem2017practical}.
However, because some of the original data cannot be discarded, there is an upper limit on the compression rate of the data selection method.
\par
Recently, dataset distillation as an alternative method to the data selection has attracted widespread attention~\cite{wang2018datasetdistillation}.
Dataset distillation is the task of synthesizing a small dataset that preserves most information of the original large dataset.
The algorithm of dataset distillation takes a sizable real dataset as the input and synthesizes a small distilled dataset.
Unlike the data selection method that uses actual data from the original dataset, dataset distillation generates synthetic data with a different distribution from the original one~\cite{dong2022privacy}.
Therefore, the dataset distillation method can distill the whole dataset into several images, or even only one image~\cite{li2022compressed}.
Dataset distillation has many application scenarios, such as privacy protection~\cite{li2020soft, li2023sharing}, continual learning~\cite{wiewel2021soft}, and neural architecture search~\cite{zhao2021datasetcondensation}, etc.
\par
Since the dataset distillation task was first introduced in 2018 by Wang et al.~\cite{wang2018datasetdistillation}, it has gained increasing attention in the research community~\cite{yu2023review}.
The original dataset distillation algorithm is based on meta-learning and optimizes distilled images by gradient-based hyperparameter optimization.
Subsequently, many studies have significantly improved distillation performance by label distillation~\cite{bohdal2020flexible}, gradient matching~\cite{zhao2021datasetcondensation}, differentiable augmentation~\cite{zhao2021differentiatble}, and distribution/feature matching~\cite{zhao2023distribution, wang2022cafe}.
The recently proposed dataset distillation method by matching network parameters has been the new state-of-the-art (SOTA) on several datasets~\cite{cazenavette2022dataset}.
However, we found that a few parameters are difficult to match during the distillation process, which degrades distillation performance.
\par
The presence of difficult-to-match parameters during dataset distillation is due to data heterogeneity. This heterogeneity arises from differences and variations in the training datasets used for the teacher and student networks. 
While the teacher network is trained on a large, original dataset, the student network is trained on a compressed distilled dataset. 
Data heterogeneity introduces discrepancies in data distribution and representation between the teacher and student datasets. 
As a result, certain patterns and critical knowledge may be underrepresented or even absent in the distilled dataset. 
Consequently, the absence of crucial information in the distilled dataset can lead to some parameters in the student network being unable to sufficiently match their corresponding counterparts in the teacher network, giving rise to the emergence of difficult-to-match parameters.
\par
In this study, we propose a new dataset distillation method using parameter pruning.
As one of the model pruning approaches, parameter pruning is frequently used for model compression and accelerated model training.
Here, we introduce parameter pruning into dataset distillation to remove the effect of difficult-to-match parameters.
The proposed method can synthesize more robust distilled datasets by pruning difficult-to-match parameters during the distillation process, improving the distillation and cross-architecture generalization performance.
Experimental results on two benchmark datasets show the superiority of the proposed method to other SOTA dataset distillation methods.
\begin{figure*}[t]
        \centering
        \includegraphics[width=14cm]{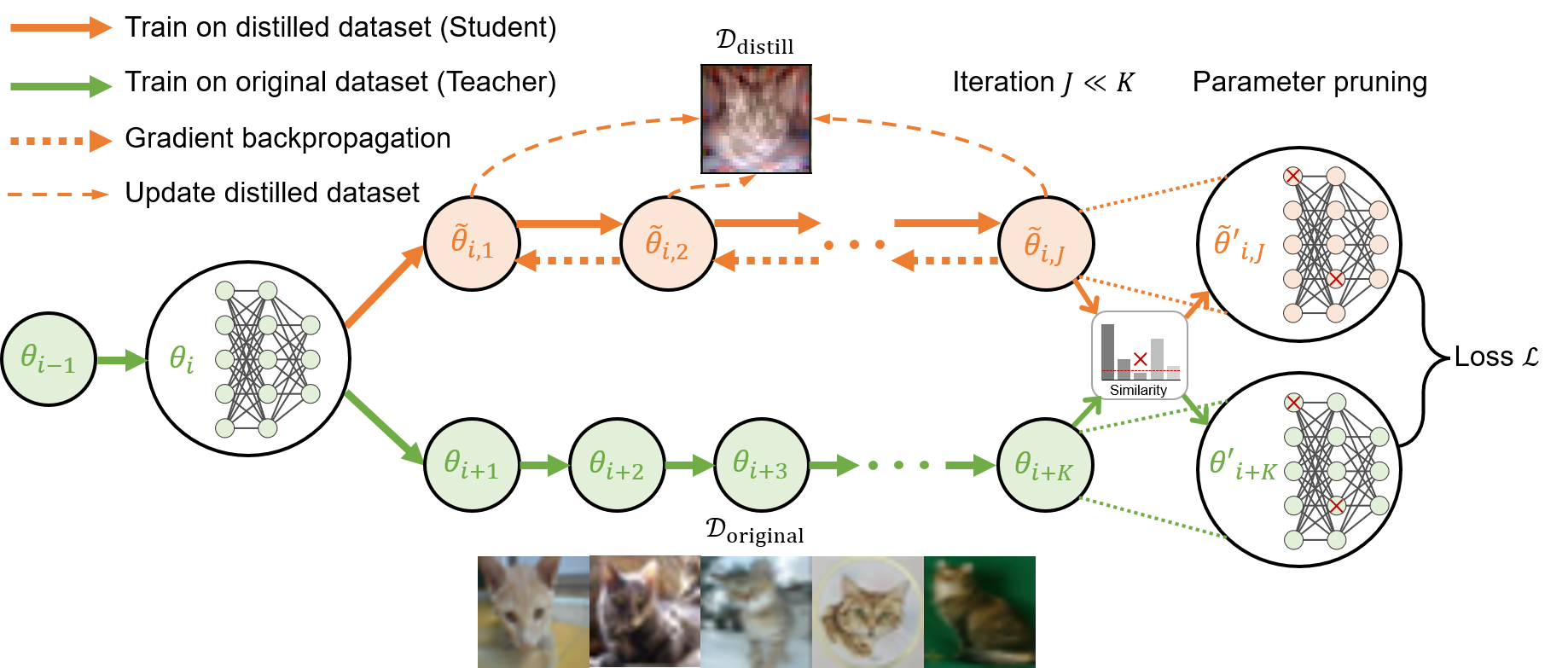}
        \caption{Overview of the proposed method. Our method uses a teacher-student architecture, and the objective is to make the student network parameters $\tilde{\theta}'_{i,J}$ match the teacher network parameters $\theta'_{i+K}$.}
        \label{fig1}
\end{figure*}
\par
Our main contributions can be summarized as follows:
\begin{itemize}
    \item We propose a new dataset distillation method based on parameter pruning, which can synthesize more robust distilled datasets and improve the distillation performance.
    \item The proposed method outperforms other SOTA dataset distillation methods on two benchmark datasets and has better cross-architecture generalization performance.
\end{itemize}
\section{Methodology}
An overview of the proposed method is depicted in Fig.~\ref{fig1}.
Our method consists of three stages: teacher-student architecture training, teacher-student parameter matching, and optimized distilled dataset generation. 
\subsection{Teacher-Student Architecture Training}
First, we pretrain $N$ teacher networks on $\mathcal{D}_\textrm{original}$ and save their snapshot parameters at each epoch.
We define teacher parameters as time sequences of parameters $\{\theta_{i}\}^{I}_{0}$.
Meanwhile, student parameters are defined as $\tilde{\theta}_{i}$, which are trained on the distilled dataset $\mathcal{D}_\textrm{distill}$ at each training step $i$.
At each distillation step, we first sample parameters from
one of the teacher parameters at a random step $i$ and use them to initialize student parameters as $\tilde{\theta}_{i}=\theta_{i}$.
We set an upper bound $I^{+}$ on the random step $i$ to ignore the less informative later parts of the teacher parameters.
The number of updates for student and teacher parameters are set as $J$ and $K$, respectively, where $J \ll K$.
For each student update $j$, we sample a minibatch $b_{i,j}$ from a distilled dataset as follows:
\begin{equation}
b_{i,j} \thicksim \mathcal{D}_\textrm{distill}.
\end{equation}
Then we perform $j$ updates on the student parameters $\tilde{\theta}$ using the cross-entropy loss $\ell$ as follows:
\begin{equation}
\tilde{\theta}_{i,j+1} = \tilde{\theta}_{i,j} - \alpha\nabla\ell(\mathcal{A}(b_{i,j});\tilde{\theta}_{i,j}),
\end{equation}
where $\alpha$ represents the trainable learning rate.
$\mathcal{A}$ represents a differentiable data augmentation module proposed in~\cite{zhao2021differentiatble}, which can improve the distillation performance.
\begin{algorithm}[t]
    \caption{Dataset Distillation Using Parameter Pruning}
    \label{alg1}
    \begin{algorithmic}[1]
    \REQUIRE 
    $\{\theta_{i}\}^{I}_{0}$: teacher parameters trained on $\mathcal{D}_\textrm{original}$;
    $\alpha_{0}$: initial value for $\alpha$;
    $\mathcal{A}$: differentiable augmentation function;
    $\epsilon$: threshold for pruning;
    $T$: number of distillation steps;
    $J$: number of updates for the student network;
    $K$: number of updates for the teacher network;
    $I^{+}$: maximum start epoch.
    \ENSURE
    optimized distilled dataset $\mathcal{D}^{\ast}_\textrm{distill}$ and
    learning rate $\alpha^{\ast}$.
    \\
    \STATE
    Initialize distilled dataset:
    $\mathcal{D}_\textrm{distill} \thicksim \mathcal{D}_\textrm{original}$
    \STATE
    Initialize trainable learning rate:
    $\alpha = \alpha_{0}$
    \FOR{each distillation step $t = 0$ to $T - 1$}
    \STATE
    Choose random start epoch $i < I^{+}$
    \STATE
    Initialize student network with teacher parameter: 
    $\tilde{\theta}_{i}=\theta_{i}$
    \FOR{each student update $j = 0$ to $J - 1$}
    \STATE
    Sample a minibatch of distilled dataset:
    $b_{i,j} \thicksim \mathcal{D}_\textrm{distill}$ 
    \STATE
    Update student network with cross-entropy loss:
    \STATE
    $\tilde{\theta}_{i,j+1} = \tilde{\theta}_{i,j} - \alpha\nabla\ell(\mathcal{A}(b_{i,j});\tilde{\theta}_{i,j})$
    \ENDFOR
    \IF{parameter similarity in $\tilde{\theta}_{i,J}$ and $\theta_{i+K}$ is less than $\epsilon$}
    \STATE
    Prune difficult-to-match parameters: Eqs. (3)--(7)
    \ENDIF
    \STATE
    Compute loss between the pruned parameters:
    \STATE
    $\mathcal{L} = || \tilde{\theta}'_{i,J}-\theta'_{i+K} ||^{2}_{2} \,\,\,/\,\,\, || \theta'_{i}-\theta'_{i+K} ||^{2}_{2}$
    \STATE
    Update $\mathcal{D}_\textrm{distill}$ and $\alpha$ with respect to $\mathcal{L}$
    \ENDFOR
    \end{algorithmic}
\end{algorithm}
\begin{table*}[t]
    \footnotesize
    \centering
    \caption{Test accuracy of different methods on CIFAR-10 and CIFAR-100.}
    \label{tab1}
    \begin{tabular}{l|c|cccccccc|cc}
    \hline
    & IPC & Random & Forgetting~\cite{toneva2019empirical} & Herding~\cite{chen2010super} & DSA~\cite{zhao2021differentiatble} & DM~\cite{zhao2023distribution} & CAFE~\cite{wang2022cafe} & MTT~\cite{cazenavette2022dataset} & Ours & Full Dataset\\\hline
    \multirow{3}*{CIFAR-10} & 1 & 14.4$\pm$2.0 & 13.5$\pm$1.2 & 21.5$\pm$1.2 & 28.8$\pm$0.7 &  26.0$\pm$0.8 & 31.6$\pm$0.8 & 46.3$\pm$0.8 & \bfseries{46.4$\pm$0.6} & \multirow{3}*{84.8$\pm$0.1} \\
    & 10 & 26.0$\pm$1.2 & 23.3$\pm$1.0 & 31.6$\pm$0.7 & 52.1$\pm$0.5 & 48.9$\pm$0.6 & 50.9$\pm$0.5 & 65.3$\pm$0.7 & \bfseries{65.5$\pm$0.3} &  \\
    & 50 & 43.4$\pm$1.0 & 23.3$\pm$1.1 & 40.4$\pm$0.6 & 60.6$\pm$0.5 & 63.0$\pm$0.4 & 62.3$\pm$0.4 & 71.6$\pm$0.2 & \bfseries{71.9$\pm$0.2} &  \\\hline
          
    \multirow{3}*{CIFAR-100} & 1 & 4.2$\pm$0.3 & 4.5$\pm$0.2 & 8.4$\pm$0.3 & 13.9$\pm$0.3 & 11.4$\pm$0.3 & 14.0$\pm$0.3 & 24.3$\pm$0.3 & \bfseries{24.6$\pm$0.1} & \multirow{3}*{56.2$\pm$0.3} \\
    & 10 & 14.6$\pm$0.5 & 15.1$\pm$0.3 & 17.3$\pm$0.3 & 32.3$\pm$0.3 & 29.7$\pm$0.3 & 31.5$\pm$0.2 & 40.1$\pm$0.4 & \bfseries{43.1$\pm$0.3} & \\
    & 50 & 30.0$\pm$0.4 & 30.5$\pm$0.3 & 33.7$\pm$0.5 & 42.8$\pm$0.4 & 43.6$\pm$0.4 & 42.9$\pm$0.2 & 47.7$\pm$0.2 & \bfseries{48.4$\pm$0.3} & \\\hline
    \end{tabular}
\end{table*}
\subsection{Teacher-Student Parameter Matching}
Next, we obtain the student parameters $\tilde{\theta}_{i,J}$ trained on the distilled dataset $\mathcal{D}_\textrm{distill}$ from $J$ updates after initializing the student network.
Meanwhile, we can obtain the teacher parameters $\theta_{i+K}$ trained on the original dataset $\mathcal{D}_\textrm{original}$ from $K$ updates, which are the known parameters that have been pretrained.
Next, we transform the student parameters $\tilde{\theta}_{i,J}$ and teacher parameters $\theta_{i+K}$ into one-dimensional vectors as follows:
\begin{equation}
\tilde{\theta}_{i,J} = [\tilde{\theta}_{i,J}^{1}, \tilde{\theta}_{i,J}^{2},\dotsb,\tilde{\theta}_{i,J}^{p}],
\end{equation}
\begin{equation}
\theta_{i+K} = [\theta_{i+K}^{1}, \theta_{i+K}^{2},\dotsb,\theta_{i+K}^{p}],
\end{equation}
where $p$ represents the total number of parameters. 
If the numerical similarity of a parameter pair $\frac{\tilde{\theta}_{i,J}^{x}}{\theta_{i+K}^{x}}$ or $\frac{\theta_{i+K}^{x}}{\tilde{\theta}_{i,J}^{x}} < \epsilon$, where $\epsilon$ is a threshold, the parameter is recognized as difficult-to-match parameter.
The index $x$ of the difficult-to-match parameter is remembered and then automatically pruned in $\tilde{\theta}_{i,J}$, $\theta_{i+K}$, and $\theta_{i}$.
The remaining effective parameters are defined as follows:
\begin{equation}
\tilde{\theta}'_{i,J} = [\tilde{\theta}_{i,J}^{1}, \tilde{\theta}_{i,J}^{2},\dotsb,\tilde{\theta}_{i,J}^{u}],
\end{equation}
\begin{equation}
\theta_{i+K}' = [\theta_{i+K}^{1}, \theta_{i+K}^{2},\dotsb,\theta_{i+K}^{u}],
\end{equation}
\begin{equation}
\theta_{i}' = [\theta_{i}^{1}, \theta_{i}^{2},\dotsb,\theta_{i}^{u}],
\end{equation}
where $u$ represents the number of remaining effective parameters.
When pruning is applied, the less important or redundant parameters are eliminated, leading to a more concise representation of the student network.
This process helps the student network align more closely with the teacher network, as it reduces the impact of data heterogeneity-induced discrepancies and improves the likelihood of parameter matching. 
By discarding irrelevant information, pruning allows the student network to focus on essential patterns and knowledge, thus mitigating the negative effects of information absence in the distilled dataset. 
Consequently, the alignment of parameter values between the teacher and student networks becomes more feasible, and the challenge of difficult-to-match parameters is alleviated.
The final loss $\mathcal{L}$ calculates the normalized squared $L_{2}$ error between the remaining effective student parameters $\tilde{\theta}'_{i,J}$ and teacher parameters $\theta'_{i+K}$ as follows:
\begin{equation}
\mathcal{L} = \frac{|| \tilde{\theta}'_{i,J}-\theta'_{i+K} ||^{2}_{2}} {|| \theta'_{i}-\theta'_{i+K} ||^{2}_{2}},
\end{equation}
where we normalize the $L_{2}$ error by the distance $\theta'_{i}-\theta'_{i+K}$ related to the teacher so that we can still obtain proper supervision from the late training period of the teacher network even if it has converged.
In addition, the normalization process eliminates cross-layer and neuronal differences in magnitude.
\subsection{Optimized Distilled Dataset Generation}
Finally, we minimize the loss $\mathcal{L}$ using momentum stochastic gradient descent and backpropagate the gradients through all $J$ updates to the student network for updating the pixels of the distilled dataset $\mathcal{D}_\textrm{distill}$ and trainable learning rate $\alpha$.
Note that the process of determining the optimized learning rate $\alpha^{\ast}$ can function as an automatic adjustment for the number of student and teacher updates (i.e., hyperparameters $J$ and $K$).
The distillation process of the proposed method is summarized in Algorithm~\ref{alg1}.
After obtaining the optimized distilled dataset $\mathcal{D}^{\ast}_\textrm{distill}$, we can train different neural networks on it for efficiency and use for downstream tasks, such as continual learning and neural architecture search.
\section{Experiments}
\subsection{Experimental Settings}
We used two benchmark datasets (i.e., CIFAR-10 and CIFAR-100) in the experiments for comparison with other methods.
The resolution of the images in CIFAR-10 and CIFAR-100 is 32 $\times$ 32.
For comparative methods, we used three data selection methods: random selection (Random), example forgetting (Forgetting)~\cite{toneva2019empirical}, and herding method (Herding)~\cite{chen2010super}.
The random selection method offers simplicity while lacking informative example prioritization. 
Example forgetting method aims to reduce redundancy, potentially capturing diverse patterns, yet risks information loss from underrepresented examples. 
The herding method focuses on uncertain examples to enhance robustness.
However, the method is computationally demanding.
\par
Furthermore, we used five SOTA dataset distillation methods: differentiable siamese augmentation (DSA)~\cite{zhao2021differentiatble}, distribution matching (DM)~\cite{zhao2023distribution}, aligning features (CAFE)~\cite{wang2022cafe}, matching training trajectories (MTT)~\cite{cazenavette2022dataset} and kernel inducing point (KIP)~\cite{nguyen2021kipimprovedresults}.
Among the SOTA dataset distillation methods, DSA employs Siamese networks for augmentation and end-to-end training. 
However, optimal hyperparameter tuning is essential for DSA.
DM aligns datasets' distributions to enhance parameter alignment and its effectiveness is influenced by the distribution-matching strategy employed.
CAFE focuses on feature-level alignment via feature matching, with efficacy dependent on feature complexity and network architecture.
MTT aligns training evolution for dynamic knowledge transfer that necessitates meticulous tuning and consideration of difficult-to-match parameters.
KIP employs kernel methods to facilitate robust knowledge transfer.
However, its effectiveness is influenced by the computational complexity it introduces and the critical decision of choosing an appropriate kernel function.
\par
The network used in this study is a sample 128-width ConvNet~\cite{gidaris2018dynamic}, which is frequently used in current dataset distillation methods.
We conducted two experiments to verify the effectiveness of the proposed method: benchmark comparison, and cross-architecture generalization.
We found that pruning too many parameters would crash model training.
Hence, the parameter pruning threshold $\epsilon$ was set to 0.1, which performed well in all experiments.
All experimental results are the average accuracy and standard deviation of five networks trained from scratch on the distilled dataset.
\subsection{Benchmark Comparison}
In this subsection, we verify the effectiveness of the proposed method by comparing it with other SOTA dataset distillation methods on two benchmark datasets: CIFAR-10 and CIFAR-100.
We employed zero-phase component analysis (ZCA) whitening with default parameters and used a 3-depth ConvNet the same as MTT~\cite{cazenavette2022dataset}.
We pretrained 200 teacher networks (50 epochs per teacher) for the distillation process.
The number of distillation steps was set to 5,000.
The number of images per class (IPC) was set to 1, 10, and 50, respectively.
For KIP~\cite{nguyen2021kipimprovedresults}, we used their original 1024-width ConvNet (KIP-1024) and 128-width ConvNet (KIP-128) for a fair comparison.
Furthermore, we used their custom ZCA implementation for distillation and evaluation.
\begin{figure}[t]
        \centering
        \includegraphics[width=8cm]{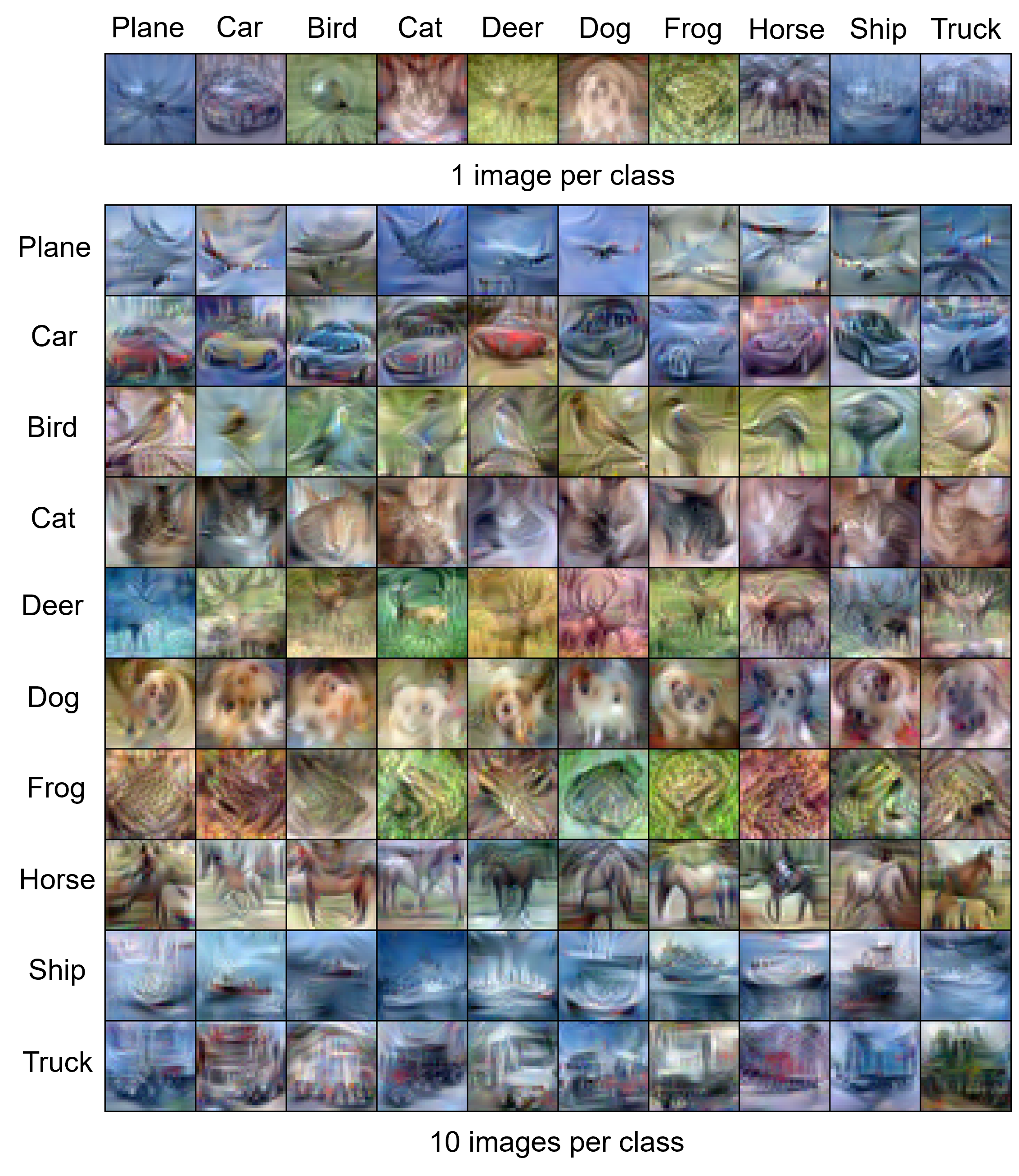}
        \caption{Visualization results of the distilled CIFAR-10 dataset.}
        \label{fig2}
\end{figure}
\begin{table}[t]
    \footnotesize
    \centering
    \caption{Test accuracy of different width KIP~\cite{nguyen2021kipimprovedresults} and our method on CIFAR-10 and CIFAR-100.}
    \label{tab2}
    \begin{tabular}{l|c|c|cc}
    \hline
    & IPC & KIP-1024 & KIP-128 & Ours-128 \\\hline
    \multirow{3}*{CIFAR-10} & 1 & 49.9 &  38.3 & \bfseries{46.4} \\
    & 10 & 62.7 & 57.6 & \bfseries{65.5} \\
    & 50 & 68.6 & 65.8 & \bfseries{71.9} \\\hline
          
    \multirow{3}*{CIFAR-100} & 1 & 15.7 & 18.2 & \bfseries{24.6} \\
    & 10 & 28.3 & 32.8 & \bfseries{43.1}\\
    & 50 & - & - & \bfseries{48.4} \\\hline
    \end{tabular}
\end{table}
\par
Table~\ref{tab1} shows that the proposed method outperformed the dataset selection methods and SOTA dataset distillation methods in all settings.
Especially for CIFAR-100 with IPC = 10, our method increased accuracy by 3.0\% compared to the second-best method MTT.
As listed in Table~\ref{tab2}, the proposed method drastically outperformed KIP using the same 128-width ConvNet.
Even for KIP that uses 1024-width ConvNet, our method has higher accuracy except for CIFAR-10 with 1 image per class.
For the results of CIFAR-100 with IPC = 50, KIP did not conduct experiments due to the large computational resources and time required; thus, we only report our results in this paper.
\par
Figure~\ref{fig2} shows the visualization results of the distilled CIFAR-10 dataset.
As depicted in Fig.~\ref{fig2}, when we set the number of distilled images to 1, the resulting images were not only more abstract but also more information-dense than the original images because all information about a class has to be compressed into only one image during the distillation process.
Meanwhile, when the number of distilled images was set to 10, the resulting images were more realistic and contained various forms because discriminative features in a class can be compressed into multiple images during the distillation process.
For example, we can see various types of dogs and different colored cars.
\begin{table}[t]
    \footnotesize
    \centering
    \caption{Cross-architecture generalization results on CIFAR-10 dataset with IPC = 10.}
    \label{tab3}
    \begin{tabular}{lcccc}
    \hline
    Architecture & ConvNet & AlexNet & VGG11 & ResNet18 \\\hline
    Ours & \bfseries{65.4$\pm$0.4} & \bfseries{35.8$\pm$1.3} & \bfseries{52.9$\pm$0.9} & \bfseries{51.8$\pm$1.1} \\
    MTT~\cite{cazenavette2022dataset} & 64.3$\pm$0.7 & 34.2$\pm$2.6 & 50.3$\pm$0.8 & 46.4$\pm$0.6 \\
    KIP~\cite{nguyen2021kipimprovedresults} & 47.6$\pm$0.9 & 24.4$\pm$3.9 & 42.1$\pm$0.4 & 36.8$\pm$1.0 \\\hline
    \end{tabular}
\end{table}
\subsection{Cross-Architecture Generalization}
In this subsection, we verify the effectiveness of our method in cross-architecture generalization.
A cross-architecture means using distilled images generated by one architecture and testing on other architectures.
The distilled images were generated by ConvNet on CIFAR-10 and the number of distilled images was set to 10.
We used the same pretrained teacher networks used in subsection 3.2 for rapid distillation and experimentation.
For KIP, we used 128-width ConvNet and their custom ZCA implementation for distillation and evaluation.
We also tested the accuracy of ConvNet and three cornerstone networks for the evaluation of cross-architecture generalization: AlexNet~\cite{krizhevsky2012imagenet}, VGG11~\cite{simonyan2015very}, and ResNet18~\cite{he2016deep}.
\par
Table~\ref{tab3} shows that our method outperformed the SOTA methods MTT and KIP for all architectures.
Especially for ResNet, our method increased accuracy by 5.2\% compared with MTT.
The results indicate that our method generated more robust distilled images than the other methods.
By pruning difficult-to-match parameters in teacher and student networks, the proposed method can avoid the influence of these parameters on the distilled dataset, improving cross-architecture generalization performance.
\section{Conclusion}
This study proposed a novel dataset distillation method based on parameter pruning.
The proposed method can synthesize more robust distilled datasets by pruning difficult-to-match parameters during the distillation process.
The experimental results show that the proposed method outperforms other SOTA dataset distillation methods on two benchmark datasets and has better cross-architecture generalization performance.
\bibliographystyle{IEEEbib}
\bibliography{refs}

\begin{thebibliography}{10}

\bibitem{liu2017survey}
Weibo Liu, Zidong Wang, Xiaohui Liu, Nianyin Zeng, Yurong Liu, and Fuad~E
  Alsaadi,
\newblock ``A survey of deep neural network architectures and their
  applications,''
\newblock {\em Neurocomputing}, vol. 234, pp. 11--26, 2017.

\bibitem{xiong2022ld}
Mingkang Xiong, Zhenghong Zhang, Tao Zhang, and Huilin Xiong,
\newblock ``Ld-net: A lightweight network for real-time self-supervised
  monocular depth estimation,''
\newblock {\em IEEE Signal Process. Lett.}, vol. 29, pp. 882--886, 2022.

\bibitem{bachem2017practical}
Olivier Bachem, Mario Lucic, and Andreas Krause,
\newblock ``Practical coreset constructions for machine learning,''
\newblock {\em arXiv:1703.06476}, 2017.

\bibitem{wang2018datasetdistillation}
Tongzhou Wang, Jun-Yan Zhu, Antonio Torralba, and Alexei~A. Efros,
\newblock ``Dataset distillation,''
\newblock {\em arXiv:1811.10959}, 2018.

\bibitem{dong2022privacy}
Tian Dong, Bo~Zhao, and Lingjuan Liu,
\newblock ``Privacy for free: How does dataset condensation help privacy?,''
\newblock in {\em Proc. Int. Conf. Mach. Learn.}, 2022, pp. 5378--5396.

\bibitem{li2022compressed}
Guang Li, Ren Togo, Takahiro Ogawa, and Miki Haseyama,
\newblock ``Compressed gastric image generation based on soft-label dataset
  distillation for medical data sharing,''
\newblock {\em Comput. Methods Programs Biomed.}, 2022.

\bibitem{li2020soft}
Guang Li, Ren Togo, Takahiro Ogawa, and Miki Haseyama,
\newblock ``Soft-label anonymous gastric x-ray image distillation,''
\newblock in {\em Proc. IEEE Int. Conf. Image Process.}, 2020, pp. 305--309.

\bibitem{li2023sharing}
Guang Li, Ren Togo, Takahiro Ogawa, and Miki Haseyama,
\newblock ``Dataset distillation for medical dataset sharing,''
\newblock in {\em Proc. AAAI Conf. Artif. Intell., Workshop}, 2023, pp. 1--6.

\bibitem{wiewel2021soft}
Felix Wiewel and Bin Yang,
\newblock ``Condensed composite memory continual learning,''
\newblock in {\em Proc. Int. Jt. Conf. Neural Netw.}, 2021, pp. 1--8.

\bibitem{zhao2021datasetcondensation}
Bo~Zhao and Hakan Bilen,
\newblock ``Dataset condensation with gradient matching,''
\newblock in {\em Proc. Int. Conf. Learn. Represent.}, 2021.

\bibitem{yu2023review}
Ruonan Yu, Songhua Liu, and Xinchao Wang,
\newblock ``A comprehensive survey to dataset distillation,''
\newblock {\em arXiv:2301.07014}, 2023.

\bibitem{bohdal2020flexible}
Ondrej Bohdal, Yongxin Yang, and Timothy Hospedales,
\newblock ``Flexible dataset distillation: Learn labels instead of images,''
\newblock in {\em Proc. Adv. Neural Inf. Process. Syst., Workshop}, 2020.

\bibitem{zhao2021differentiatble}
Bo~Zhao and Hakan Bilen,
\newblock ``Dataset condensation with differentiable siamese augmentation,''
\newblock in {\em Proc. Int. Conf. Mach. Learn.}, 2021, pp. 12674--12685.

\bibitem{zhao2023distribution}
Bo~Zhao and Hakan Bilen,
\newblock ``Dataset condensation with distribution matching,''
\newblock in {\em Proc. IEEE/CVF Wint. Conf. Appl. Comput. Vision}, 2023.

\bibitem{wang2022cafe}
Kai Wang, Bo~Zhao, Xiangyu Peng, Zheng Zhu, Shuo Yang, Shuo Wang, Guan Huang,
  Hakan Bilen, Xinchao Wang, and Yang You,
\newblock ``{CAFE}: Learning to condense dataset by aligning features,''
\newblock in {\em Proc. IEEE/CVF Conf. Comput. Vision Pattern Recognit.}, 2022,
  pp. 12196--12205.

\bibitem{cazenavette2022dataset}
George Cazenavette, Tongzhou Wang, Antonio Torralba, Alexei~A. Efros, and
  Jun-Yan Zhu,
\newblock ``Dataset distillation by matching training trajectories,''
\newblock in {\em Proc. IEEE/CVF Conf. Comput. Vision Pattern Recognit.}, 2022,
  pp. 4750--4759.

\bibitem{toneva2019empirical}
Mariya Toneva, Alessandro Sordoni, Remi Tachet~des Combes, Adam Trischler,
  Yoshua Bengio, and Geoffrey~J Gordon,
\newblock ``An empirical study of example forgetting during deep neural network
  learning,''
\newblock in {\em Proc. Int. Conf. Learn. Represent.}, 2019.

\bibitem{chen2010super}
Yutian Chen, Max Welling, and Alex Smola,
\newblock ``Super-samples from kernel herding,''
\newblock in {\em Proc. Conf. Uncertainty Artif. Intell.}, 2010.

\bibitem{nguyen2021kipimprovedresults}
Timothy Nguyen, Roman Novak, Lechao Xiao, and Jaehoon Lee,
\newblock ``Dataset distillation with infinitely wide convolutional networks,''
\newblock in {\em Proc. Adv. Neural Inf. Process. Syst.}, 2021, pp. 5186--5198.

\bibitem{gidaris2018dynamic}
Spyros Gidaris and Nikos Komodakis,
\newblock ``Dynamic few-shot visual learning without forgetting,''
\newblock in {\em Proc. IEEE/CVF Conf. Comput. Vision Pattern Recognit.}, 2018,
  pp. 4367--4375.

\bibitem{krizhevsky2012imagenet}
Alex Krizhevsky, Ilya Sutskever, and Geoffrey~E Hinton,
\newblock ``Imagenet classification with deep convolutional neural networks,''
\newblock in {\em Proc. Adv. Neural Inf. Process. Syst.}, 2012, pp. 1097--1105.

\bibitem{simonyan2015very}
Karen Simonyan and Andrew Zisserman,
\newblock ``Very deep convolutional networks for large-scale image
  recognition,''
\newblock {\em Proc. Int. Conf. Learn. Represent.}, 2015.

\bibitem{he2016deep}
Kaiming He, Xiangyu Zhang, Shaoqing Ren, and Jian Sun,
\newblock ``Deep residual learning for image recognition,''
\newblock in {\em Proc. IEEE/CVF Conf. Comput. Vision Pattern Recognit.}, 2016,
  pp. 770--778.

\end{thebibliography}

\end{document}